\documentclass[preprint,12pt]{elsarticle}

\usepackage{amssymb}
\usepackage{amsmath}

\usepackage[nodots]{numcompress}

\usepackage{hyperref}

\usepackage{multirow}

\journal{arXiv: Computer Sciences}

\begin{document}

\begin{frontmatter}



\title{Combining patch-based strategies and non-rigid registration-based label fusion methods}


\author{Carlos Platero, M.Carmen Tobar}

\address{Health Science Technology Group, Technical University of Madrid, Ronda de Valencia 3, 28012, Madrid, Spain. }

\begin{abstract}
The objective of this study is to develop a patch-based labeling method that cooperates with a label fusion using non-rigid registrations. We present a novel patch-based label
fusion method, whose selected patches and their weights are calculated
from a combination of similarity measures between patches using
intensity-based distances and labeling-based distances, where a
previous labeling of the target image is inferred through a label fusion
method using non-rigid registrations. These combined similarity
measures result in better selection of the patches, and their weights
are more robust, which improves the segmentation results compared to
other label fusion methods, including the conventional patch-based labeling method.
To evaluate the performance and the robustness of the proposed label fusion method, we employ two available databases of
T1-weighted (T1W) magnetic resonance imaging (MRI) of human brains. We compare our approach with other label fusion methods in the automatic hippocampal segmentation from T1W-MRI.

Our label fusion method yields mean Dice coefficients of 0.847 and 0.798 for the two databases used with mean times of approximately 180 and 320 seconds, respectively. The collaboration between the patch-based labeling method and the label fusion using non-rigid registrations is given in the several levels: (a) The pre-selection of the patches in the atlases are improved, (b) The weights of our selected patches are also more robust, (c) our approach imposes geometrical restrictions, such as shape priors, and (d) the work-flow is very efficient. We show that the proposed approach is very competitive with respect to recently reported methods.
\end{abstract}

\begin{keyword}

Atlas-based segmentation \sep Image registration \sep Patch-based label fusion \sep Hippocampal segmentation \sep Magnetic resonance imaging
\end{keyword}

\end{frontmatter}





%
\section{Introduction}
Magnetic resonance imaging (MRI) plays a crucial role in
quantitatively measuring the differences in anatomical structures between either individuals or groups.
In many clinical studies, the hippocampal volumetry from MRI is one of the bio-markers used to examine the diagnosis of Alzheimer disease (AD)~\cite{leung2010automated,nestor2013direct}. Volumetric hippocampal measurements are the best predictors of AD conversion in subjects with mild cognitive impairment~\cite{clerx2013measurements}.

Manual volumetry is considered the gold standard, but it is time consuming. Consequently, many automatic approaches have been proposed to extract hippocampal structures from brain MRI. Among them, atlas-based methods have been demonstrated to outperform other algorithms~\cite{babalola2009evaluation}. In the context of this study, an atlas is an image in one modality with its respective labeling (typically generated by manual segmentation)~\cite{aljabar2009multi}. Atlas-based segmentation is motivated by the observation that segmentation strongly correlates with the context information. After warping the atlas to the target image, context is directly transferred from the atlas to the target image. In medical images, context plays a very important role because the anatomical structures are primarily constrained to relatively fixed positions.

However, the disease status of the subject-atlases used in this
approach may affect the quality of the results. Therefore, atlases should be customized for pathological studies. There are atlases in epilepsy~\cite{jafari2011dataset} and in AD~\cite{frisoni2014eadc} for hippocampal segmentation, which capture the significant morphological variations that occur in both disease processes.

Assuming that manual labeling is the ground truth, the errors produced
by atlas-based segmentation can be primarily attributed to the
registration step. Segmentations with a single atlas are intrinsically
biased toward the shape and the appearance of a
subject~\cite{barnes2008comparison}. Several studies have shown that
approaches that incorporate the properties of a group of atlases
outperform the use of a single
atlas~\cite{aljabar2009multi,rohlfing2005quo,heckemann2006automatic,lotjonen2010fast,collins2010towards}. Multi-atlas
segmentation is a popular approach for labeling anatomical structures
from medical images. A subset of atlases within a database are
registered to a target image, and their segmentations can be
transformed and subsequently fused to provide a consensus
segmentation. The main benefit of the multi-atlas label fusion is that
the effect of errors associated with any single atlas propagation can
be reduced in the process of combination. The main drawback of this
approach is its computational complexity. Indeed, the computational
time for segmentation increases linearly with the number of atlases
that have to be registered. Although the availability and low cost of
multi-core processors are making this approach more feasible, an atlas selection is generally required such that the number of atlases is as low as possible so that no further improvement is expected when adding more atlases~\cite{aljabar2009multi,klein2008automatic,van2010adaptive}. Furthermore, a label fusion method is also required to obtain the resulting segmentation. This paper is focused on the label fusion problem.

The label fusion methods have been classified into two categories: global weighted voting and local weighted voting. Most existing label fusion methods are based on global weighted voting, such as majority voting~\cite{rohlfing2005quo} (MV), STAPLE~\cite{warfield2004simultaneous} or weighted voting
(WV)~\cite{artaechevarria2009combination}, which are widely used in
medical image segmentation. In these approaches, each atlas
contributes to the resulting segmentation with the same weight for all
of its voxels. It is very sensitive to registration errors because it does not take into account the relevance of each sample. Recent works have shown that local weighted voting methods outperform global methods~\cite{artaechevarria2009combination,sabuncu2010generative,coupe2011patch}.

Moreover, the errors differ if the registration is affine or non-rigid. In affine transformations, the structures of the target image and the atlases are only aligned in position, orientation and scale, but the context information of the aligned atlas is preserved in the domain of the target image. However, the atlases are typically warped to the target image using non-rigid registration techniques. This approach presents the advantage of forcing the resulting segmentation to have a similar global shape to those of expert-labeled structures in the atlases.
In these approaches, there is a one-to-one mapping between the target image and each atlas.
The major drawback of applying non-rigid registration to the atlases
is the smoothing of the transferred labels. Indeed, because of the
regularization constraints involved in these registrations, some
details can be lost, and the local high variability cannot be
captured. We propose combining the information from the registered
atlases by means of an affine as non-rigid registrations.

In atlas-warping by non-rigid registrations, label fusion methods
generally calculate the labels associated with the target image via
maximum a posteriori (MAP)
estimation~\cite{lotjonen2010fast,sabuncu2010generative,van2008hippocampus}. A
statistical model is constructed from the registered atlases, which is
normally decomposed into an appearance model and a shape prior
term. Appearance models are able to combine many of the local
intensity statistics. These models have difficulty in taking the
regional information into account. For this reason, appearance models
are joined with shape prior models. The MAP inference can be obtained
via expectation
maximization~\cite{sabuncu2010generative,ledig2012multi} or by transforming the labeling problem in terms of energy minimization~\cite{lotjonen2010fast,van2008hippocampus,song2006integrated,wolz2010measurement,platero2014multiatlas}.

Patch-based label fusion methods have shown great potential using only
affine
registrations~\cite{coupe2011patch,rousseau2011supervised,wang2013multi,tong2013segmentation}. Rather
than fusing label maps as in multi-atlas segmentation, this framework
is based on the non-local mean principle~\cite{Buades06}. Each voxel
is represented by a small image patch.  Similar patches on each atlas
image are aggregated to a reference voxel in the target image based on
the non-local mean. The more similar a patch of a voxel in an atlas
image is to a reference voxel in the target image, the higher is the
weight that is used to propagate its labeling to the reference voxel
in the target image. All selected patches from a subset of atlases
with their weighed labelings are fused to estimate the labeling of the
reference voxel. These methods exhibit two interesting properties: (i)
this approach drastically increases the number of samples considered
during the labeling estimation, and (ii) the local intensity context
(i.e., patch) can be used to produce a robust comparison among
samples~\cite{coupe2011patch}. Therefore, the usual assumption of
one-to-one mapping in the label fusion using non-rigid registrations
is relaxed by using local search windows. Although these methods are
powerful, they highly depend on the intensity-based similarity
measures between patches. For selected patches with similar appearance according to the similarity measures used, their corresponding labelings may be very different. Image similarities over small image patches may not be an optimal estimator~\cite{wang2013multi}. Moreover, the labeling is local and independent, without global constraints. To overcome these drawbacks, Asman and Landman~\cite{asman2013non} have proposed an iterative algorithm between the estimation of the labeling and the weights of the patches that maximizes the expected value of a conditional likelihood function. Alternatively, we propose an algorithm without iterations. An estimate of the labeling with constraints in the shape priors is inferred by a label fusion method based on non-rigid registrations. Then, a patch-based labeling method is applied, which selects the patches, and their weights are calculated using similarity measures of the intensities and binary labelings.
These combined similarity measures produce better selection of the patches, and their weights are more robust, which improves the segmentation results compared to other existing approaches.

In this paper, brain MRI are used to validate the proposed
framework. These images generally show different structures of
interest to be segmented. Therefore, a region-wise approach is more
appropriate~\cite{han2007atlas}, which can be achieved by dividing the
image into multiple anatomically meaningful
regions~\cite{shi2010construction}. Once the regions of interest
(ROIs) are defined, a ranking of atlases is
calculated for each ROI~\cite{aljabar2009multi}, and the selected registered
atlases are fused into each ROI of the target image.  Partitioning the
problem in ROIs improves the registration and segmentation
results. Indeed, the multi-atlas approaches have greater accuracy when the registrations are only made near the object of interest and not in the entire image~\cite{shi2010construction}. Furthermore, these approaches convert the complex multi-label problem into feasible binary segmentation problems.

Finally, we test different label fusion methods on publicly available
MRI of human brains. We demonstrate that our approach produces
as good or even better automatic segmentations than other label fusion methods.
\\

The remainder of this paper is organized as follows. In Section 2, the proposed method of combining a patch-based labeling method with an atlas-warping using non-rigid registrations is presented. The experiments and results for the hippocampal segmentation are described in Section 3. The discussion and conclusions are presented in Section 4.

%
\section{Methods}
We propose a patch-based labeling method that cooperates with
atlas-warping using non-rigid registrations. First, a subset of $N_R$
atlases are registered non-rigidly into the target image, and then a
label fusion method is applied. The label fusion method is based on
minimizing a pseudo-Boolean function using graph cuts with information
of appearance, shape and
context~\cite{lotjonen2010fast,van2008hippocampus,song2006integrated,wolz2010measurement,platero2015crf}. Then,
a patch-based labeling method is applied using the above segmentation
of the target image. The patches of another subset of $N_A$ atlases,
which are registered by affine transformations, are pre-selected with
a structural similarity measure that takes into account both the
intensity and the labeling of the candidate patches. From the selected
patches, a multi-point label estimation is calculated for each voxel
that belongs to the target image. The weights of the selected patches are computed from a combination of $L2$-norm measures between patches using intensity-based distances and labeling-based distances. The following paragraphs explain the methods used and how they work together.

\subsection{Label fusion using atlas-warping by non-rigid registrations}\label{subSection:CRFModel}
Given a ROI of the target image $I$, a subset of $N_R$ atlases
$\{A_i\}_{i=1, \dots , N_R} =\{I_i,S_i\}_{i=1, \dots , N_R}$ are used
for the label fusion method using non-rigid registrations, where
$I_i:\Omega_i \subset \mathbb{N}^3 \rightarrow \mathbb{R}$ are the
modality images and $S_i:\Omega_i \subset \mathbb{N}^3 \rightarrow
\{0,1\}$ are the label maps. In the labeled images, voxels that belong
to the anatomical structure of interest are designated by the label
$S(x)=1$ and background voxels are designated by the label
$S(x)=0$. We denote $\Phi_i:\Omega \rightarrow \Omega_i$ as the
spatial mapping from the target image coordinates to the coordinates
of the $i-$th atlas. For simplicity, we assume that $\{\Phi_i\}_{i=1,
  \dots , N_R}$ have been pre-computed using a pairwise registration
procedure. This assumption allows us to shorthand
$\mathbb{A}=\{\tilde{S}_i= S_i \circ \Phi_i, \tilde{I}_i = I_i \circ
\Phi_i\}_{i=1,\dots,N_R}$ as the atlases into the coordinates of the
target image. We seek to minimize an energy function under the
Bayesian formulation, which defines the conditional probability as a
discrete random field $S$ with a neighborhood system
$\mathcal{E}$. The neighborhood system $\mathcal{E}$ is the set of
edges that connect variables in the random field. A conditional random field model~\cite{lafferty2001conditional} (CRF) is defined by the following pseudo-Boolean function:

\begin{equation}\label{eq:CRFModel}
  E(S) = \sum_{x \in \Omega} \psi_x (S(x);\theta_1(I,\mathbb{A})) + \lambda \sum_{x,y \in \mathcal{E}} \psi_{xy} (S(x),S(y);\theta_2(I)),
\end{equation}

\noindent where $\Theta=\{\theta_1,\theta_2\}$ are the model parameters for this ROI and $\lambda$ is a tunable parameter that determines the trade-off between
the unary and pairwise potentials. We next define the form of the two potential functions and their parameters.

\subsubsection{The unary potentials}\label{subsection:unary}
The unary potentials $\psi_x (S(x);\theta_1(I,\mathbb{A}))$ use the
Bayesian formulation, which allows prior information about the shape
and appearance of structures to be segmented to be incorporated in the
model. 

\subsubsection*{Image likelihood}
We assume that the observed intensities of $I$ are independent random variables. The image likelihood $p(I|S;\mathbb{A})$ can then be written as a product of the likelihoods of the individual voxels:

$$ p(I|S;\mathbb{A})=\prod_{x \in \Omega}p(I(x)|S(x);\mathbb{A}).$$

A discriminative appearance model with low computational effort is selected. The registered atlas images are convolved with a filter-bank. A set of feature extraction kernels $\alpha_j(x)$ are used to produce different feature maps:

$$F_{\tilde{I}_i}(x)=\{\tilde{I}_i(x) * \alpha_j(x) \}_{j=1,\ldots,d},$$

\noindent where $d$ is the dimension of the feature vector and $F_{\tilde{I}_i}(x)$ denotes the resulting feature vector of $\tilde{I}_i$ at voxel $x$ associated with the filter $\{\alpha_j(x)\}_{j=1,\ldots,d}$. In this paper, derivatives of Gaussians~\cite{leung2001representing,shotton2009textonboost} and 3D steerable filters~\cite{derpanis2005three} are adopted to extract features (for further details, see section \ref{subsec:DS}). The responses for all registered atlas image voxels are whitened separately (to provide zero mean and unit covariance). These feature vectors are used to train a $k$-nearest neighbor ($k$-NN) appearance model. For computational efficiency, we use the $k$d-tree algorithm~\cite{Mount2010} to perform the nearest neighbor search. A $k$d-tree model is constructed with $\{F_{\tilde{I}_i}(x),\tilde{S}_i(x)\}_{i=1,\ldots,N_R, x \in  \Omega}$. The target image is also convolved and whitened.

For each label $l \in \{0,1\}$, we consider
$\mathcal{F}_l=\{F_{\tilde{I}_i}(x)\:/\: \tilde{S}_i(x) =l\}$ as the
set of feature vectors extracted from the voxels that belong to the registered atlas images and whose labels are $l$. Let $\{F_r\}_{r\in R_l(x)}\subset \mathcal{F}_l$ be the set whose elements are nearest neighbors to $F_I(x)$ and $R_l(x)$ be the set of the indices of the feature vectors with label $l$ that are nearest neighbors to
$F_I(x)$. The image likelihoods of the individual voxels that belong to the target image are calculated using the following formula:

\begin{equation}\label{eq:DiscModel}
p(I(x)|l;\mathbb{A}) \propto \sum_{r\in R_l}\exp\left(-\|F_I(x)-F_r\|^2_2\right).
\end{equation}

\subsubsection*{Label prior}
The label prior probability $ p(S;\mathbb{A},I)$ models the joint probability of all voxels that belong to the ROI in a particular label configuration. However, we assume that the prior probability that voxel $x$ has label $l$ only depends on its position, the similarity between $I$ and $\tilde{I}_i$ and the transferred atlas-labeled images:

$$ p(S;I,\mathbb{A})=\prod_{x \in \Omega}p(S(x);I,\mathbb{A}).$$

\noindent This assumption is not realistic, but we encode the correlations of the labels using pairwise potentials. For each voxel $x$ and each label $l \in \{0,1\}$, we define:
$$u(S(x)=l;I,\mathbb{A})=\sum_{i \in Q_l(x)} m(I,\tilde{I}_{i},x)^q,$$

\noindent where $Q_l(x)=\{i|\tilde{S}_i(x)=l\}$, $m(I,\tilde{I}_{i},x)$ is a local or global similarity measure between the target image and the registered atlas image at $x$, and $q$ is an associated gain exponent~\cite{artaechevarria2009combination}. The prior probability is defined as

\begin{equation}\label{eq_WV}
p(S(x)=l;I,\mathbb{A})=\frac{u(S(x)=l;I,\mathbb{A})}{\sum_{j\in\{0,1\}}u(S(x)=j;I,\mathbb{A})}.
\end{equation}
\noindent

The image likelihood and label prior terms are combined to define the unary potentials $\psi_x (S(x);\theta_1(I,\mathbb{A}))$:
$$\psi_x(S(x);\theta_1(I,\mathbb{A}))=-\log\left(\frac{p(I(x)|S(x);\mathbb{A})p(S(x);I,\mathbb{A})}{p(I(x);\mathbb{A})}\right).$$

\subsubsection{Spatial regularization}
Following the work of Song et al.~\cite{song2006integrated}, a
smoothness term is added to the energy function. These authors
combined intensity and local boundary information into the pairwise
potentials, which have been successfully applied for brain
segmentation. These pairwise potentials take the form of a contrast-sensitive Potts model:
$$\psi^R_{xy} (S(x),S(y);\theta_2(I)) = \left\{ \begin{array}{lll}
                                  0 & & \text{if} \ \ S(x) = S(y), \\
                                  \beta(S(x),S(y);\theta_2(I)) & & \text{otherwise}.
                                \end{array} \right.$$
\noindent where
\begin{align}\label{eq:Song}
\beta(S(x),S(y);\theta_2(I)) &= c\left(1+\ln\left(1+\frac{\| I(x) -I(y)\|^2}{2\sigma^2}\right)\right) \nonumber \\
&\quad + (1-c)\left(\max_{r \in M_{xy}} g(\|\nabla I(r) \|)\right). \nonumber \\
\end{align}

\noindent where $g(\|\nabla I(x) \|)=1- \exp\left(-\frac{\|\nabla I(x)
  \|}{\sigma_G}\right)$ with a normalization factor
$\sigma_G$. $M_{xy}$ is a line that joins $x$ and $y$, and $\sigma$ is the robust scale of image $I$. The parameter $0 \leq c \leq 1$ controls the influence of the
boundary-based and intensity-based parts.

\subsection{Patch-based labeling method}
Given a subset of the aligned atlases in a ROI of the target image
using affine transformations, each voxel $x$ in $I$ is correlated to
each voxel $y$ in $\tilde{I}_i$ with a weight of $\omega_i(x,y)$.  In the conventional method, the weights are defined by intensity-based similarity measures between patches. A patch-based representation is extracted using a sub-volume centered at $x$ from the target image or the aligned atlas. These signatures are denoted by $P_I(x)$ and $\{P_{\tilde{I}_i}(y),P_{\tilde{S}_i}(y)\}$, respectively. The signature difference between a reference voxel $x$ and a voxel in an atlas image $y$ is used to define $\omega_i(x,y)$:

\begin{equation}\label{eq:ConvPacth}
\omega_i(x,y) = \left\{ \begin{array}{ll}
                               e^{-\frac{\|P_I(x)-P_{\tilde{I}_i}(y)\|^2_2}{h^2(x)}} & \text{if} \,\,\,y \in \mathcal{N}(x), \\
                               0 & \text{otherwise}, \\
                             \end{array}
\right.
\end{equation}

\noindent where $\mathcal{N}(x)$ denotes the neighborhood of voxel $x$
in the atlas images. The parameter $h$ determines how this method
assigns the weights for each patch $P_{\tilde{I}_i}(y)$. We use $h$ to
show the minimal distance between the target patch $P_I(x)$ and its neighboring patches $P_{\tilde{I}_i}(y)$~\cite{coupe2011patch}.

Using these weights, we can use a single-point model for the label fusion from the atlases, as follows:

$$ S(x)= \frac{\sum^{N_A}_{i=1} \sum_{y \in \Omega} \omega_i(x,y) \tilde{S}_i(y)}{\sum^{N_A}_{i=1} \sum_{y \in \Omega} \omega_i(x,y)}, $$

\noindent where $S(x)$ is an estimation of label '1' for every voxel $x$ and $N_A$ is the number of the selected aligned atlases used to fuse. However, because the similarity measure is based on patches, one can obtain a multi-point model for the label fusion~\cite{rousseau2011supervised}:

$$ P_S(x)= \frac{\sum^{N_A}_{i=1} \sum_{y \in \Omega} \omega_i(x,y) P_{\tilde{S}_i}(y)}{\sum^{N_A}_{i=1} \sum_{y \in \Omega} \omega_i(x,y)}, $$

\noindent where $P_{\tilde{S}_i}(y)$ is a label patch that belongs to
the $i-$th atlas and $P_S(x)$ is a label patch estimation of the
target image at $x$, i.e., for each voxel, a vector of the likelihood for label '1' is calculated. These estimates are then aggregated using a combination classifier. In this work, we used the majority voting rule to fuse these estimates. The label patch estimation provides better results than the single estimation~\cite{rousseau2011supervised}.

\subsection{Combining patch-based and registration-based label fusion}

The similarity measure is the core of the patch-based labeling methods, and this measure is only calculated using intensity-based distances. The selected patches from the atlases can have similar distances to the patch of the reference voxel, but nevertheless, their label patches may be very different. In addition, these approaches estimate the labels in a local manner, without global constraints such as the shape priors. To overcome this drawback, a new similarity measure based on the labeling distance is added.

We consider the label patch as a clique, i.e., a connected set of voxels, whose labels are conditionally dependent on each other. A label patch estimation can be inferred by minimizing (\ref{eq:CRFModel}) using global shape constraints. Next, we add a measure using label-based distances:

\begin{equation}\label{eq:MixPacth}
\omega_i(x,y) = \left\{ \begin{array}{ll}
                               e^{-\frac{\|P_I(x)-P_{\tilde{I}_i}(y)\|^2_2}{h^2_I(x)}}e^{-\frac{\|P_S(x)-P_{\tilde{S}_i}(y)\|^2_2}{h^2_S(x)}} & \text{if} \,\,\,y \in \mathcal{N}(x), \\
                               0 & \text{otherwise}. \\
                             \end{array}
\right.
\end{equation}

\noindent where $h_I(x)$ and $h_S(x)$ are defined as the minimal distances between the target patch and its neighboring patches according to the intensities and labelings, respectively (for further details, see section \ref{subsec:patch}). Now, the weights of the patches are calculated considering both the local appearance and the global shape. Moreover, this proposal does not require the addition of more tunable parameters compared to the path-based conventional labeling method.

%
\section{Experiments with Brain MRI data}
To evaluate the performance and the robustness of the proposed label
fusion method, we select hippocampal segmentation because it is one
of the most studied problems in the analysis of brain images. We
employ two available databases of T1-weighted (T1W) MRI: (i) 18
modified images from the Internet Brain Segmentation Repository
(IBSR)~\cite{IBSRweb,rohlfing2012image} and (ii) 50 images of
epileptic and nonepileptic patients with hippocampal outlines
(HFH)~\cite{jafari2011dataset}. IBSR contains images of healthy
patients with expert segmentation of 43 anatomical structures. The
voxel spacing of these images is 0.9375 x 1.5 x 0.9375 $mm^3$. In
contrast, HFH contains a total of 50 images, which were randomly
divided into 25 images used for the training set and the other 25 for
the test set. Manual segmentations are only available for the training
images. Images were acquired using two MRI systems with
different field strengths (1.5 T and 3.0 T) and with different resolutions (0.78 x 2 x 0.78 $mm^3$ and 0.39 x 2 x 0.39 $mm^3$).

The proposed segmentation scheme involves four main steps: (1) MRI pre-processing, (2) spatial normalization and definition of the ROIs, (3) the first labeling based on atlas warping using non-rigid registrations and (4) the final labeling by patches based on similarity measures in intensity and labeling.

During pre-processing of the databases, non-brain regions are removed from all structural images. Removing non-brain tissue prior to registration is generally accepted as a means to simplify the inter-subject registration problem and thus increase the quality of the registrations~\cite{battaglini2008enhanced,klein2009evaluation}. The images are skull-stripped using BET~\cite{smith2002fast}. Then, all images are spatially normalized to a reference atlas using an affine registration. For the IBSR database, CMTK's affine registration tool is used~\cite{studholme1999overlap}. In HFH, an atlas (HFH\_021) is selected as a reference to which all atlases are then co-registered with an affine transformation using FLIRT with 12 degrees of freedom~\cite{jenkinson2002improved}. After spatial normalization for both of the databases, a region of interest is defined for each structure studied (left and right hippocampus) as the minimum bounding box containing the structure for all of the training atlases expanded by three voxels along each dimension. The patient image is also processed using a skull-striping filter, and an affine transformation is applied to the common reference space. Then, the normalized patient image is cropped around the structures of interest.

For each ROI, the normalized atlas images are first ranked based on their similarity according to the normalized target image using the mutual information (MI) measure~\cite{viola1997alignment}. Then, the $N_R$-selected atlases are registered non-rigidly into the ROI of the normalized target image. Next, the registered atlases are fused, and the labeling is calculated using graph cuts based on minimizing the energy function of (\ref{eq:CRFModel}).

\begin{figure}[h]
\centering
\includegraphics[width=\textwidth,height=6cm]{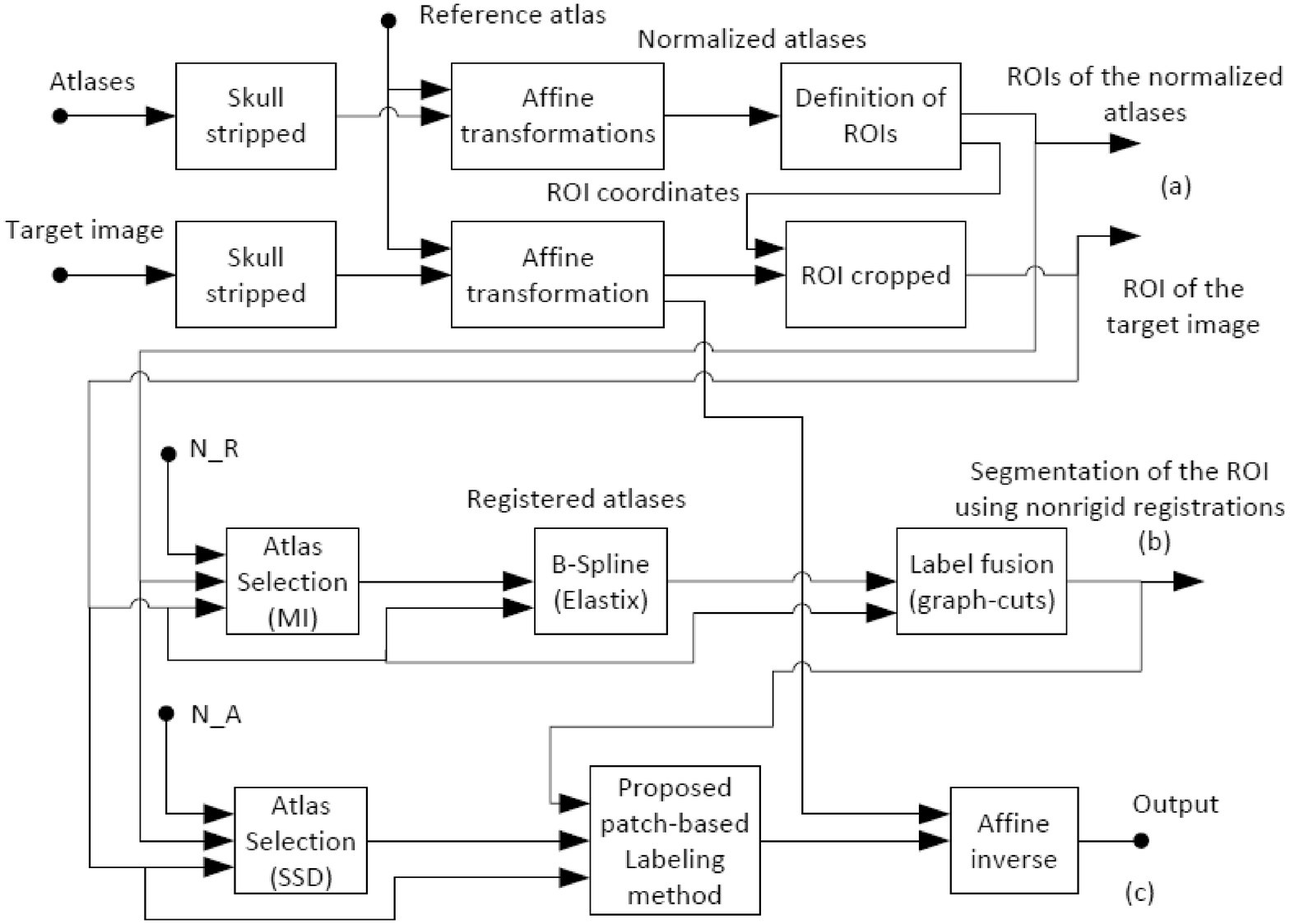}
\caption{Flow chart summarizing the processing of the patient images: (a) MRI pre-processing, spatial normalization and definition of the ROIs. (b) Segmenting of the ROIs using the non-rigid registration-based label fusion method. (c) Segmenting of the ROIs using the proposed patch-based labeling method.}
\label{fig:flowchar_hippo}
\end{figure}

Next, an intensity normalization is applied to the above normalized
atlas images using the histogram matching
algorithm~\cite{gonzales6digital} with the ROI of the normalized
target image as a reference. Now, the sum of the squared difference (SSD) measure between each atlas image and the target image is used to rank and select the first $N_A$ atlases. This measure is chosen because SSD is related to the similarity between patches in intensities. Finally, from the previous labeling of the target image and the $N_A$-selected atlases, the segmentation result is obtained using the proposed patch-based labeling method, and an inverse affine transformation is applied to return the automatic segmentation into the native space of the target image. Fig.~\ref{fig:flowchar_hippo} shows a flow chart that summarizes the processing of the images.

Although in the theoretical framework (see section \ref{subSection:CRFModel}), the registrations were performed from the atlases into the target image, there is no loss of generality using the spatial normalization by using affine transformations. This intermediate step improves the computational efficiency. Furthermore, the ranking of the atlases for both fusion methods is immediately calculated with the spatial normalization. The normalized atlases are also used to obtain the library of the patches for implementing the patch-based labeling method.

\subsection{Experiments with the training atlases}
For each database, a leave-one-out validation strategy is performed to
determinate the tunable parameters. These parameters are varied in
certain ranges, and their effects are measured from the overlap
between the resulting segmentation and the ground truth. The DICE
coefficient~\cite{dice1945measures} is chosen as the measure of the segmentation overlaps: $\text{DICE}(X,Y)=\frac{2 |X \cap Y|}{|X|+|Y|}$, where $X$ and $Y$ represent the automatic and  manual segmentation
binary images, respectively. The function $|\cdot|$ indicates the size of '1' in the set. In atlas warping by non-rigid registrations, the parameters $\{\Theta_1,\Theta_2\}$ in (\ref{eq:CRFModel}) of each ROI are first learned by piecewise training and then recombined with the tunable weight $\lambda$. With respect to the patch-based labeling method, the influence of the patch pre-selection process, the patch and search volume size, the local adaptation of $h$ and the number of the fused atlases are studied.

\subsubsection{Setting the parameters of the label fusion method using non-rigid registrations}\label{subsec:DS}
The non-rigid registrations, the atlas selection and the label fusion method have to be investigated to improve the performance of the multi-atlas segmentation approach. For each ROI, the atlases are ranked based on their similarity according to the target image using MI~\cite{aljabar2009multi}. Once the atlases are ranked, we employ a leave-one-out validation strategy to determinate the number of atlases $N_R$ that are fused to the target image. In our experiments, the first $15$ atlases more similar to the target image were registered non-rigidly ($N_R=15$)~\cite{platero2015crf}. Then, the selected atlases are co-registered non-rigidly to the ROI of the target image. All non-rigid registrations are computed using \emph{Elastix}~\cite{klein2010elastix}, a publicly available package for medical image registration. Non-rigid registration of images is based on the maximization of MI, in combination with a deformation field parameterized by cubic B-splines~\cite{rueckert1999nonrigid}.  The MI is implemented according to~\cite{thevenaz2000optimization}, using a joint histogram size of 32 x 32 and cubic B-spline Parzen windows. A unique resolution is employed using a B-spline control point spacing of 3.0 mm in all directions. To optimize the cost function, an iterative stochastic gradient descent optimizer is used~\cite{klein2007evaluation}. In each iteration, 2000 random samples are used to calculate the derivative of the cost function. A maximum of 500 iterations of the stochastic optimization procedure is used. The above-described settings were determined through trial-and-error experiments on two image pairs. These parameters of the non-rigid registrations are equally applicable to both databases.

The atlas-labeled images are modeled using the logarithm of odds (LogOdds) formulation, which is based on the signed distance transform~\cite{pohl2006logarithm}. This representation replaces the labels by the signed distances, which are assumed to be positive inside the structure of interest. We find that the LogOdds model produces more accurate results compared with trilinear interpolation or nearest-neighbor interpolation for transferring the atlas-labeled images~\cite{sabuncu2010generative}.

Then, the registered atlases are fused, and the labeling is calculated based on minimizing the energy function of (\ref{eq:CRFModel}). We now examine the impact of the label fusion parameters using atlas-warping by non-rigid registrations. Regarding the discriminative appearance model, we have explored three types of feature vectors: derivatives of Gaussians, 3D steerable filters and patches. The filter-banks provided similar results to the patches. The feature vectors using filter-banks are applied to the images because their dimensions $d$ are smaller than patches.

The experimental results lead us to choose a 12-dimensional filter
bank, which is applied in the HFH database. The images are convolved
with Gaussians at scales of $1$, $2$ and $4$; derivatives of Gaussians
at scales of $2$ and $4$; and Laplacians of Gaussians at scales of
$1$, $2$ and $4$~\cite{shotton2009textonboost}. Following our
experiments, in the IBSR database, a 16-dimensional steerable filter
bank is used~\cite{derpanis2005three}: the 6 basis functions of the
second derivatives of Gaussian and 10 basic functions of the Hilbert
transform. We hypothesize that the difference in the filter bank used in each database is due to the voxel spacing. The Gaussian-based filter bank is better if the voxel spacing is strongly anisotropic.

We choose a k-NN appearance model due to the compromise between  performing the estimation and computational efficiency. This model could be replaced by other discriminative approaches, such as sophisticated randomized trees or boosting-based classifiers. Preliminary results on random forests~\cite{breiman2001random} have shown similar results but with a higher computational cost (note that the discriminative model must also be trained at runtime). From the proposed k-NN appearance model and given the feature vector of a voxel belonging to the target image, the training vectors in the $k$d-tree that are nearest are found. These vectors are used to calculate the distances to each label, and then equation (\ref{eq:DiscModel}) is applied for obtaining $p(I(x)|l;\mathbb{A})$. The amount of training data in the discriminative model is often biased toward the background class. A classifier learned on these data will have a prior preference for this class. To normalize for this bias, we weight each training example by the inverse class frequency. The classifiers trained using this weighting tend to provide better performance~\cite{shotton2008semantic}.

In the weighted voting method for estimating the label prior
probabilities, similarity measures are required between regions of the
target image and each registered atlas image, i.e.,
$m(I,\tilde{I}_i,x)$ in (\ref{eq_WV}).  A semi-global strategy is used
to calculate the weight for each registered atlas. This strategy is
most appropriate when the contrast between neighboring structures is
low, as in the case of the
hippocampus~\cite{artaechevarria2009combination}. A binary mask is
used for measuring this similarity between the target image and the
registered atlases. This mask is constructed by joining all
transferred labeled images. A voxel is considered in the binary mask
if at least one vote of the foreground class is received. Because a
statistical relationship is assumed among the intensities of these
images, MI is used as a similarity measure. The gain exponent is set to $q=4$~\cite{artaechevarria2009combination}.

Finally, we consider a 3D grid-graph with 26 neighborhood systems in
$\mathcal{E}$, and $\nabla I(x)$ is calculated from the derivatives of
Gaussians at scale $1$ for obtaining the pairwise potentials in
(\ref{eq:Song}). The final trained parameters for both database were
$\lambda = 0.2$, $c = 0.6$ (IBSR), and $c = 0.8$ (HFH).

\subsubsection{Setting the parameters of the patch-based labeling method}\label{subsec:patch}
\begin{figure}[h]
\centering
\includegraphics[width=\textwidth,height=6cm]{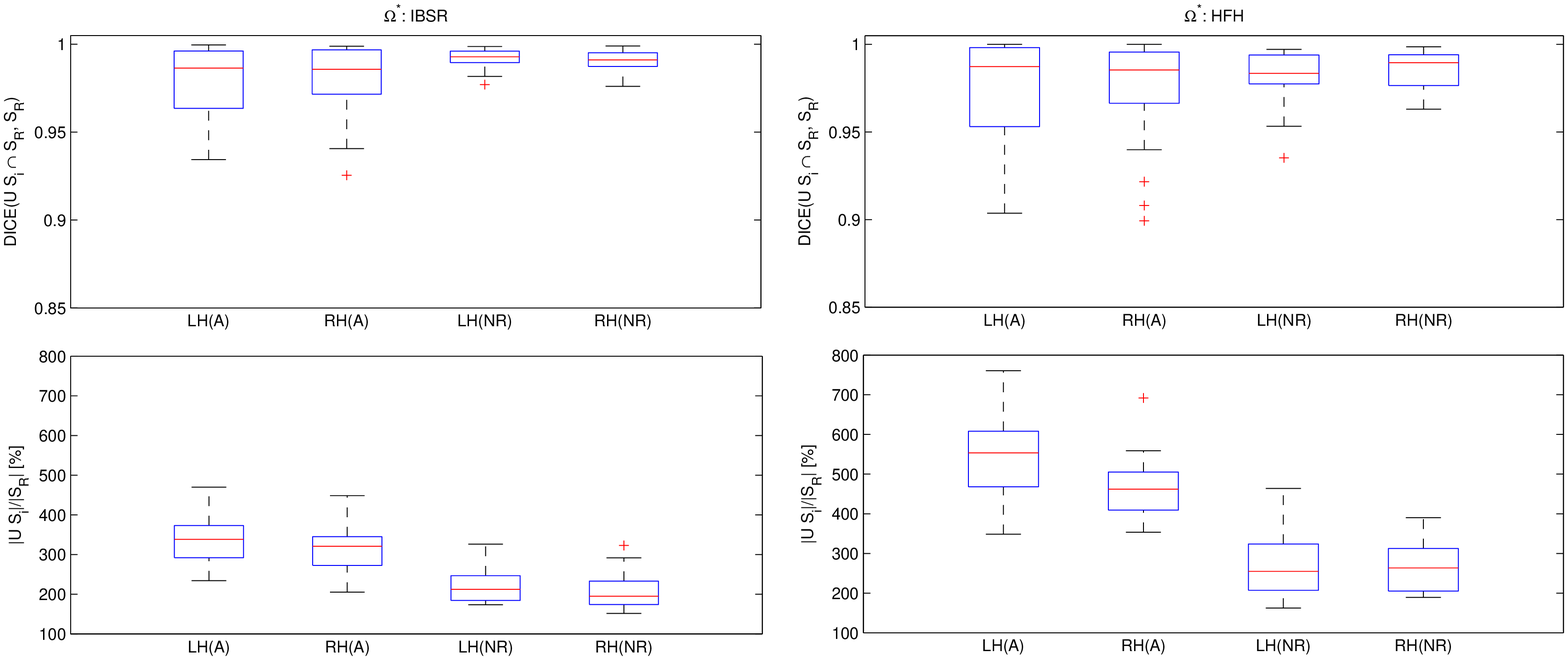}
\caption{Comparison of $\Omega^*$ between affine transformations (A)
  and non-rigid registrations (NR) for the IBSR and HFH datasets. For
  the left (LH) and right (RH) hippocampus, the first row plots the distribution of $DICE(\bigcup \tilde{S}_i \cap S_R ,S_R)$, where $S_R$ is the ground-truth segmentation. The second row plots the distribution of $\frac{| \bigcup \tilde{S}_i |}{|S_R|} 100 [\%] $.}
\label{fig:CompDomain}
\end{figure}

A subset of the target image domain $\Omega^* \subset \Omega$ is
considered to accelerate the patch-based labeling
method~\cite{coupe2011patch}. A binary mask is given with the union of
all transferred labeled images using non-rigid registrations,
$\Omega^*= \bigcup supp(\tilde{S}_i)$. Figure~\ref{fig:CompDomain}
shows the first advantage of the collaboration between the
atlas-warping by non-rigid registrations and the patch-based labeling
method. When the domains $\Omega^*$ are calculated using non-rigid
registrations, rather than using affine transformations
as~\cite{coupe2011patch}, the overlap between $\bigcup \tilde{S}_i$
and the ground-truth segmentation $(S_R)$ is increased (i.e.,
$DICE(\bigcup \tilde{S}_i \cap S_R ,S_R)$ is higher) and the volume of
$|\bigcup \tilde{S}_i|$ decreases with respect to the manually labeled
hippocampal volumes (i.e., $\frac{| \bigcup \tilde{S}_i |}{|S_R|} 100
[\%]$ is lower).  In this experiment, two conclusions were drawn: a)
because $\Omega^*$ obtained by the non-rigid registrations are more
robust than those obtained by affine transformations, the success rates of the labelings are increased, and b) the computational time of the patch-based labeling method is reduced by having fewer voxels with uncertainty in their labels.

An atlas selection is also used to identify a subset of the normalized atlases to the normalized target image by affine transformations. First, the image intensities are normalized through histogram matching. Then, the SSD measure in $\Omega^*$ between each atlas image and the target image is used to rank the atlases.  A compromise between the performance of the patch-based labeling method and computational efficiency is to choose the first 10 atlases that most resemble the target image~\cite{rousseau2011supervised,tong2013segmentation} ($N_A=10$).

Next, for each voxel in $\Omega^*$ and whose label is uncertain, a
patch pre-selection is used to accelerate the label fusion procedure
and to improve the robustness of label fusion by excluding the
unrelated patches~\cite{coupe2011patch,wu2014generative}. The
pre-selection process uses a modified version of the well-know
structural similarity measure~\cite{wang2004image}. We consider a
similarity measure that takes both the intensity and the labeling of
the candidate patch into account:

\begin{align}
ss(x,y,i) &= \frac{4 \mu_I(x) \mu_{\tilde{I}_i}(y) \sigma_I(x)\sigma_{\tilde{I}_i}(y)}{\left(\mu^2_I(x)+\mu^2_{\tilde{I}_i}(y)\right)\left(\sigma^2_I(x)+\sigma^2_{\tilde{I}_i}(y)\right)}  \nonumber \\
&\quad \cdot \frac{4 \mu_S(x) \mu_{\tilde{S}_i}(y) \sigma_S(x)\sigma_{\tilde{S}_i}(y)}{\left(\mu^2_S(x)+\mu^2_{\tilde{S}_i}(y)\right)\left(\sigma^2_S(x)+\sigma^2_{\tilde{S}_i}(y)\right)}
\end{align}

\noindent where $\left(\mu_I(x),\sigma_I(x)\right)$ and
$\left(\mu_S(x),\sigma_S(x)\right)$ are the mean and standard
deviation of the target patches $P_I(x)$ and $P_S(x)$,
respectively. Similarly,
$\left(\mu_{\tilde{I}_i}(y),\sigma_{\tilde{I}_i}(y)\right)$ and
$\left(\mu_{\tilde{S}_i}(y),\sigma_{\tilde{S}_i}(y)\right)$ are the
mean and standard deviation of the candidate patch
$P_{\tilde{I}_i}(y)$ and $P_{\tilde{S}_i}(y)$ belonging to the $i-$th selected atlas.

\begin{figure}[h]
\centering
\includegraphics[width=\textwidth,height=6cm]{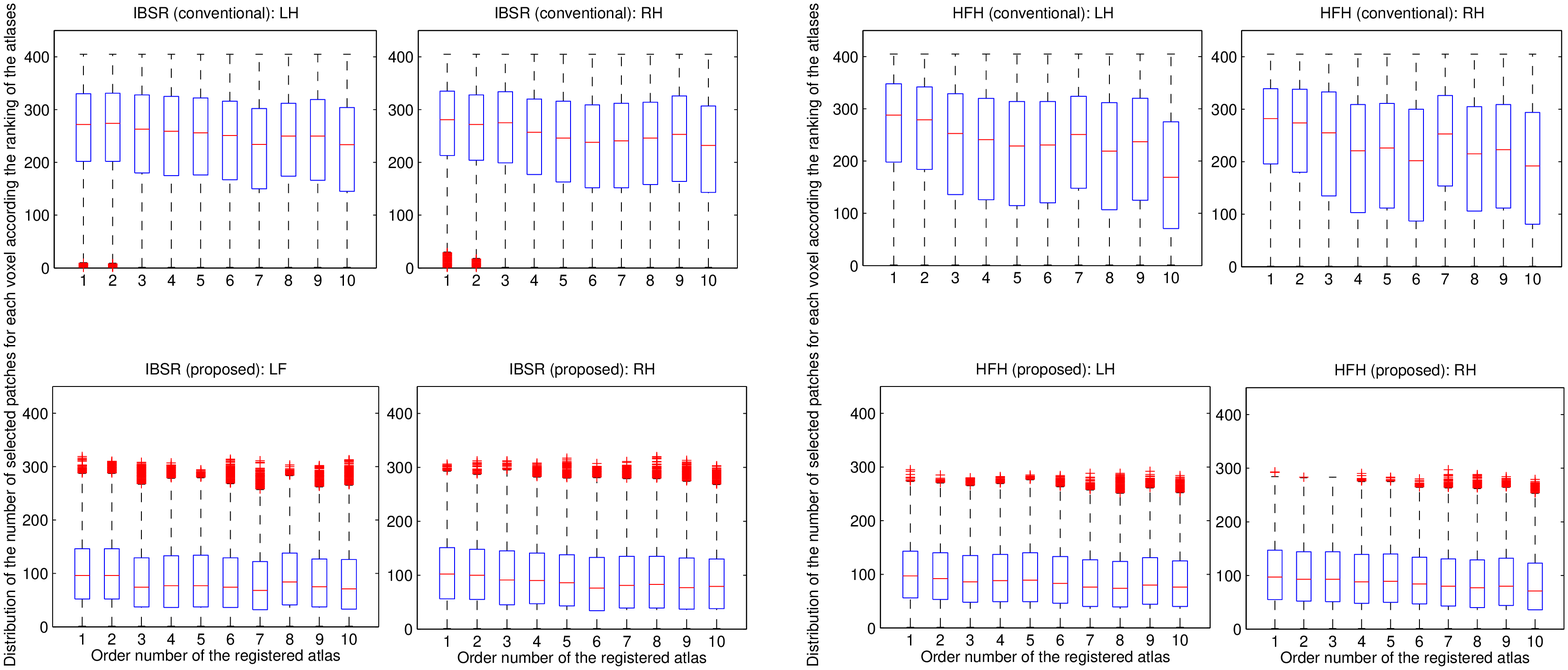}
\caption{Comparison between the conventional and proposed patch
  pre-selection procedures. The figure shows the distributions of the number of the selected patches for each voxel according to the ranking of the aligned atlases. We use $\epsilon = 0.85$.}
\label{fig:PatchDistrib}
\end{figure}

In the conventional patch-based labeling method, if the value of a
similarity measure is larger than a given threshold $\epsilon$, the
candidate patch is selected~\cite{coupe2011patch}. To avoid adding
more tuning parameters in our proposed approach, we consider a
candidate patch if our similarity measure is greater than
$\epsilon^2$, $ss(x,y,i) > \epsilon^2$. We evaluated the label fusion
accuracy when applying different thresholds during the
pre-selection. The value of $\epsilon$ was varied from $0.8$ to
$0.95$, showing no significant improvement in the label fusion
accuracy. In all experiments, we set the similarity threshold as
$\epsilon^2 = 0.85^2$. On overage, thousands of patches of the
selected aligned atlases are used to calculate the non-local mean
label fusion in every voxel. Figure~\ref{fig:PatchDistrib} shows the
distributions of the number of selected patches according to the
ranking of the aligned atlases. A comparison between the conventional
and proposed patch pre-selection procedures is shown. The number of
selected patches with our similarity measure is approximately
one-third less than that of the conventional method. Our proposal selects more
patches than the conventional pre-selection process and also improves
the robustness of label fusion by excluding the patches whose
labelings are not similar to the labeling of the target voxel. Our
approach allows geometrical constraints, such as shape priors, to be
imposed due to the labeling obtained by the atlas warping using non-rigid registrations. Indeed, the selected patches show similarity in appearance and labeling according to the target voxel. This is the second advantage of the collaboration between the atlas-warping by non-rigid registrations and the patch-based labeling method.

The impacts of the patch size and the search volume size must be investigated. The patch size is related to the complexity of the anatomical structure, and the search volume size reflects the anatomy variability of the structure of interest~\cite{coupe2011patch}. These parameters are normally defined with a $(2r +1) \times (2r +1) \times (2r +1)$ cube-shaped neighborhood by the radius $r$. In the case of the hippocampal segmentation, Coupe et al.~\cite{coupe2011patch} have reported a patch size of $7 \times 7 \times 7$ voxels and a search volume of $9 \times 9 \times 9$. Wang et al.~\cite{wang2013multi} and Tong et al.~\cite{tong2013segmentation} have proposed a patch size of $5 \times 5 \times 5$ and a search volume size of $7 \times 7 \times 7$, whereas Rousseau et al.~\cite{rousseau2011supervised} have used a patch size of $3 \times 3 \times 3$ and a search volume size of $11 \times 11 \times 11$.

\begin{figure}[h]
\centering
\includegraphics[width=\textwidth,height=6cm]{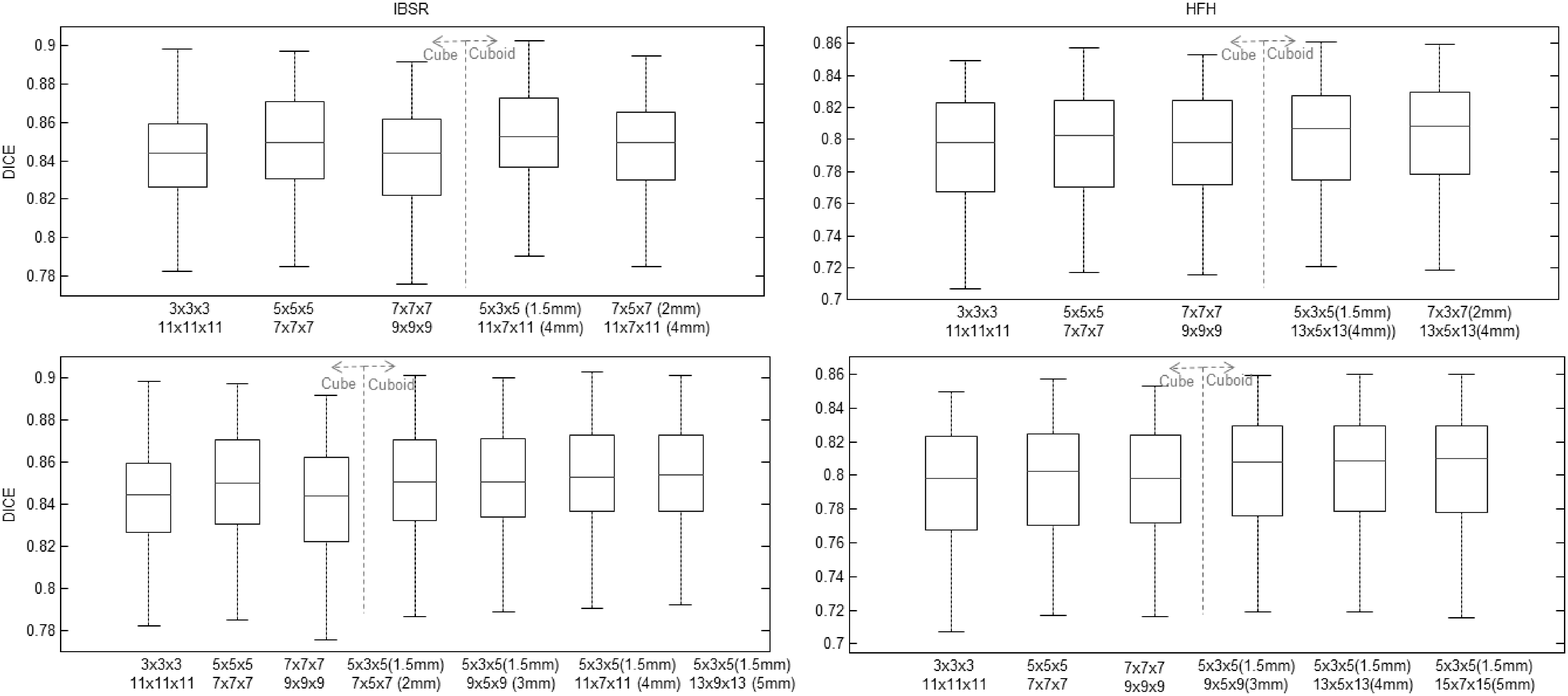}
\caption{Effects of the patch size and the search volume size on
  segmentation accuracy. The results are shown using the DICE
  coefficient distributions for both datasets. A comparison between
  the cube-shaped patch and the cuboid-shaped patch is shown. The
  first row shows the impact of the patch size with a constant search
  volume: $r_p = 1.5$ or $2$ mm in the path size with a search volume
  size of $r_s = 4$ mm. The second row shows the impact of search
  volume size with a constant path size: $r_s = 3, 4$ or $5$ mm in the
  search volume size with a path size of $r_p = 1.5$ mm. The standard cube-shaped patches are also plotted and compared with our proposal. In HFH data, images are first subsampled by a factor of two in the $X-Z$ plane to reduce the computation time. Preliminary experiments showed that using the full-resolution data increased the computation times and negligibly improved the results.}
\label{fig:rpps}
\end{figure}

Because the voxel spacing of brain T1W-MRI is not the same in
all components (particularly in the HFH database), we analyze this
parameter by taking into account the values in each direction. Now, we use a cuboid-shaped patch, which is expressed the radius $r$ in millimeters, and the patch size and the search volume size are defined by

$$\prod_{v \in \{\text{Row Spacing, Column Spacing, Slice Spacing}\}}2 \cdot \text{ceil}(r/v) +1,$$

\noindent where $\text{ceil}(x)$ rounds $x$ to toward plus infinity of the nearest integer. Figure~\ref{fig:rpps} shows the DICE distributions over varying patch and search volume sizes on both datasets. The last three above-standard cube-shaped patches are used. By contrast, the cuboid-shaped patch is defined by $r_p = 1.5$ or $2$ mm in the path size and $r_s = 3, 4$ or $5$ mm in the search volume size.  The comparison between the cube-shaped patch and the cuboid-shaped patch shows improvement when the patch size and the search volume size consider the voxel spacing.

The parameter $h$ of (\ref{eq:ConvPacth}) plays a crucial role in the
weighting of the selected patches. When $h$ is low, only a few samples
are taken into account. If $h$ is high, all selected patches tend to
have the same weight, and the estimation is similar to a classical
average. We use a modified version of $h$~\cite{coupe2011patch}, which
estimates $h$ based on the minimal distance between the target patch
$P_I(x)$ and its neighboring patches
$P_{\tilde{I}_i}(y)$~\cite{coupe2011patch}. A smoothing parameter
$\beta$ is added to the estimation of $h$, which depends on the noise level. The parameter $\beta$ is common in image denoising tasks with the non-local mean principle~\cite{coupe2008optimized} or in the kernel used by Rousseau et al.~\cite{rousseau2011supervised} in their patch-based labeling method. For low levels of noise in images, the best value of $\beta$ is close to $0.5$. For high levels of noise, this value is $1$~\cite{coupe2008optimized}. Considering the above, we define $h_I$ and $h_S$ of (\ref{eq:MixPacth}) as:

\begin{eqnarray}\label{eq:local_h}
h_I^2(x)= \beta_I \cdot \min_{y \in \mathcal{N}(x), i=1,\ldots,N_A} \left(\|P_I(x)-P_{\tilde{I}_i}(y)\|^2_2\right) + \varepsilon_I \nonumber \\
h_S^2(x)= \beta_S \cdot \min_{y \in \mathcal{N}(x), i=1,\ldots,N_A} \left(\|P_S(x)-P_{\tilde{S}_i}(y)\|^2_2\right) + \varepsilon_S,
\end{eqnarray}

\noindent where $\varepsilon_I$ and $\varepsilon_S$ are small constants to ensure numerical stability in case the patch under consideration is contained in its neighboring patches.  The value of $\beta_I$ is set to $0.5$ and $1$ for $\beta_S$. The aligned atlas images are considered to have low noise, but not the results of the label fusion method using non-rigid registrations, i.e., the similarity measures between labelings are considered to have a high noise level to relax restrictions imposed by the labeling obtained by the atlas warping.

\begin{figure}[h]
\centering
\includegraphics[width=\textwidth,height=6cm]{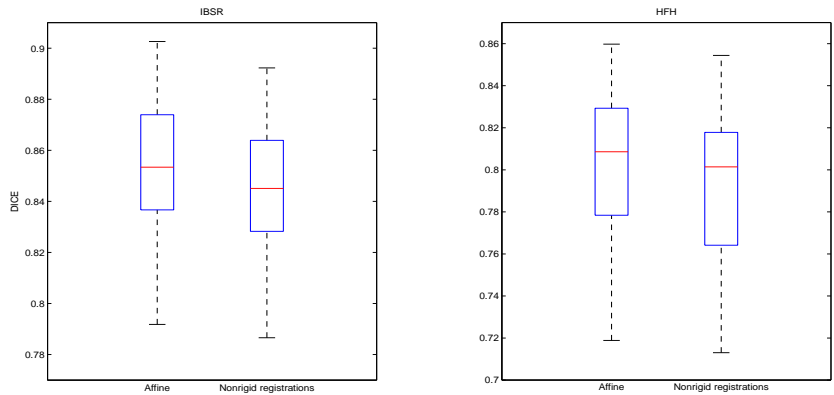}
\caption{A comparison of the proposed patch-based labeling method
  between registered atlases by affine transformations and non-rigid registrations, which are used to obtain the library of the patches.}
\label{fig:compAffineRNR}
\end{figure}

Because the registered atlases using non-rigid registrations to the
target image are available, we explore the possibility of replacing
the affine transformations by non-rigid registrations for obtaining
the library of the training patches. Figure \ref{fig:compAffineRNR}
shows that the results with the DICE distributions in both datasets
are worse using the non-rigid registrations than those using affine
transformations. We believe that the reason for this result is that the context information is better represented using patches with affine transformations than non-rigid registrations. Moreover, the proposed combination increases the robustness using information that is less correlated because the algorithm employs both aligned atlases as the registered atlases.

\subsection{Results and comparisons}
We compare our results with those obtained other label fusion
methods. Five label fusion methods are chosen: STAPLE, majority voting
(MV) and the three methods that we have used from our framework (the
conventional patch-based labeling
method~\cite{rousseau2011supervised}, the non-rigid registration-based
label fusion method and the proposed combined method). For each of the
methods in the experiments, we report the DICE coefficient as a
segmentation quality measure. Table~\ref{table:Comp_labelfusion} shows
the quantitative segmentation results for each label fusion method and
each ROI in the IBSR and HFH databases. As previously reported in~\cite{artaechevarria2009combination,rousseau2011supervised}, STAPLE does not necessarily lead to higher DICE coefficients compared to the majority voting rule.

\begin{table}
\begin{center}
\caption{Average values and standard deviations of the DICE
  coefficient for all images belonging to the IBSR and HFH databases using a) STAPLE, b) majority voting rule, c) the conventional patch-based labeling method, d) The non-rigid registration-based label fusion method and e) the proposed method. }
\label{table:Comp_labelfusion}
\begin{tabular}{|lc|c|c|}
  \hline
  Method   & ROI& IBSR & HFH \\
  \hline
  STAPLE   & LH & $0.793 \pm 0.040$ & $0.726 \pm 0.120$ \\
           & RH & $0.804 \pm 0.057$ & $0.742 \pm 0.063$ \\
  \hline
  Majority & LH & $0.791 \pm 0.040$ & $0.714 \pm 0.127$ \\
  voting   & RH & $0.798 \pm 0.052$ & $0.739 \pm 0.074$ \\
  \hline
  Patch    & LH & $0.817 \pm 0.044$ & $0.731 \pm 0.099$ \\
  labeling & RH & $0.834 \pm 0.038$ & $0.750 \pm 0.068$ \\
  \hline
  Atlas Warping   & LH & $0.833 \pm 0.042$ & $0.781 \pm 0.066$ \\
  using NR & RH & $0.841 \pm 0.049$ & $0.796 \pm 0.036$ \\
  \hline
  Proposed    & LH & $0.843 \pm 0.044$ & $0.795 \pm 0.063$ \\
  combination & RH & $0.850 \pm 0.047$ & $0.802 \pm 0.034$ \\
  \hline

\end{tabular}
\end{center}
\end{table}

Statistical significance is evaluated using the Wilcoxon signed-rank
test, where a $p-$value of $< 0.05 $ shows significant
improvement. Given the DICE coefficient distributions of the proposed
method as references, the $p-$values are shown in
Table~\ref{table:pValues} for the DICE coefficient distributions
corresponding to the other label fusion methods. These values indicate
significant improvement between our approach and other conventional
approaches. Additionally, note that there is no significant improvement when the proposed method is replaced by the atlas warping using non-rigid registrations.

\begin{table}
\begin{center}
\caption{$p-$values using the DICE coefficient distributions of the
  proposed method as a reference: a) STAPLE, b) majority voting rule, c) the conventional patch-based labeling method, and d) the atlas warping using non-rigid registrations.}
\label{table:pValues}
\begin{tabular}{|lc|c|c|}
\hline
   Approach & ROI & IBSR & HFH  \\
\hline
   STAPLE & LH & $0.0002$  & $0.002$\\
          & RH & $0.003$  &  $0.0001$\\
   \hline
   Majority & LH & $0.0003$  & $0.0001$\\
   voting   & RH & $0.0008$  & $0.0001$\\
   \hline
   Patch & LH & $0.05$  & $0.005$\\
   labeling& RH & $0.24$  & $0.0001$\\
   \hline
   Atlas Warping  & LH  & $0.45$  & $0.53$\\
   using NR& RH  & $0.42$  & $0.47$\\
   \hline
\end{tabular}
\end{center}
\end{table}

The evaluation of the test images that belong to the HFH database is performed by an external team by submitting the results to a web site~\cite{jafari2011dataset}. The given evaluations are of the entire hippocampus (see Table \ref{table:testHFH}). These results are consistent with the values obtained in the training images of the HFH database.

\begin{table}
\begin{center}
\caption{Average values and standard deviations of the DICE
  coefficient for all 25 test images belonging to the HFH database using the proposed model and the atlas warping using non-rigid registrations.}
\label{table:testHFH}
\begin{tabular}{|l|c|c|}
\hline
   Approach & ROI & HFH  \\
\hline
   Atlas Warping using NR& LH+RH & $0.778 \pm 0.047$  \\
   \hline
   Proposed Method & LH+RH & $0.788 \pm 0.044$ \\
   \hline
\end{tabular}
\end{center}
\end{table}

\begin{figure}[h]
\centering
\includegraphics[width=\textwidth,height=3cm]{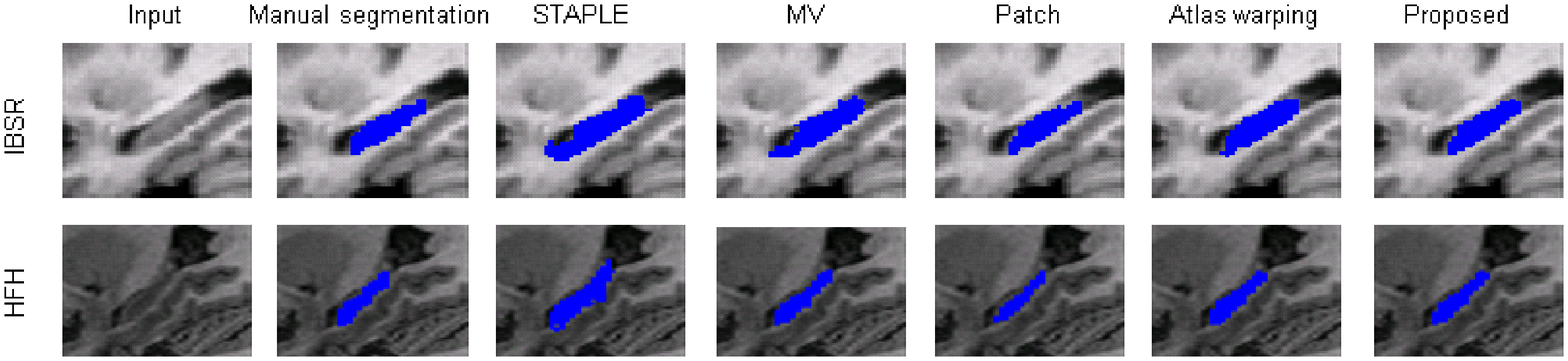}
\caption{Comparison with STAPLE, majority voting, conventional patch-based method, atlas warping using non-rigid registration and the proposed method.}
\label{fig:SegmResults}
\end{figure}

Figure~\ref{fig:SegmResults} shows the segmentation results of the
five label fusion methods for one typical subject. These qualitative
results also indicate improvement over the conventional
methods. However, there are no significant differences between the
proposed method and the atlas deformation using non-rigid registrations, as indicated by the test of statistical significance.

Table~\ref{table:compTimes} presents the average values and standard
deviations of the computing times in seconds for all training
images. We only report the times for non-rigid registrations of the
images and label fusions. The computational complexity is primarily
due to the non-rigid registrations of the selected atlases into the
target image. The computational time for segmentation increases
linearly with the number of atlases that have to be
registered. However, due to the availability and low cost of
multi-core processors, this approach is becoming more feasible. The
task of non-rigid registrations has been parallelized. The
registration of the first 15 atlases in a ROI requires less than 45
seconds. The computing times of the registration tasks have a weak
dependence on the image resolution because B-Spline uses an isotropic
grid with the same physical units (i.e., the spacing is specified in
millimeters). By contrast, the computational cost of the label fusion
method depends on the image resolution. The label fusion method using
non-rigid registrations calculates the unary and pairwise potentials
of the CRF model in (\ref{eq:CRFModel}), and the labeling is found by
applying the min-cut/max-flow algorithm
of~\cite{boykov2004experimental}. The CRF model is only trained for
voxels whose labels have uncertainty such that the computational
burden is reduced. A voxel is uncertain in its label when the
atlas-labeled images are transferred and this voxel receives votes
from different classes. Using the k-NN discriminative model with
$k$d-tree, the semi-global strategy to infer the label prior
probabilities and the graph cut techniques applied only in voxels with
uncertainty makes the implementation very efficient, even without the
optimized code. In the patch-based labeling method, many optimizations
can be used because each voxel is treated independently, which allows
multithreading. The optimized implementation should also be
investigated~\cite{coupe2008optimized}. In our first version, we
parallelize the labeling for each voxel of the target image with
uncertainty. The values of the computing times for our proposed method
include both the cost of the labeling method based on graph cuts and patches. Our approach takes an average of approximately $180$ and $320$ seconds for fusing labels (included non-rigid registrations) of both the left and right hippocampus on images belonging to IBSR and HFH, respectively. The code of the label fusion methods is not yet optimized; thus, the computing times can be easily reduced. The scripts used in this study are available at \url{https://www.nitrc.org/projects/lf_patches/}.

\begin{table}
\begin{center}
\caption{Average values and standard deviations of the computing times
  in seconds for all training images ([Dual CPU] Intel Xeon E5520 @
  2.27 GHz)}
\label{table:compTimes}
\small 
\vspace{10pt}
\begin{tabular}{|ll|c|c|} \hline
   \multirow{2}{2cm}{\bf Type} & &\multirow{2}{1.2cm}{\bf IBSR }&\multirow{2}{1.1cm}{\bf  HFH}  \\ &&&\\
\hline\hline

   \multirow{2}{2cm}{Registration}& {LH}    & $30.18 \pm 1.74$  & $41.63 \pm 0.61$\\
   & {RH}                & $31.82 \pm 1.99$  & $39.28 \pm 0.69$\\
   \hline
   {Atlas warping}& {LH}           & $13.68 \pm 2.86$  & $30.65 \pm 6.50$\\
   {using NR}& {RH}          & $13.32 \pm 2.26$  & $29.24 \pm 5.67$ \\
   \hline
   {Proposed}& {LH}           & $61.47 \pm 6.78$  & $121.66 \pm 13.69$\\
   {combination}& {RH}          & $61.43 \pm 6.87$  & $116.06 \pm 12.23$ \\
   \hline

\end{tabular}
\end{center}
\end{table}

Comparing segmentation results between different published methods is
always difficult. The quality of the databases used for validation,
the anatomical definition of the structure, the quality of expert
segmentations, the populations studied and the different measures
reported all make comparing results difficult. With these caveats in mind, we compare our segmentation results with other approaches that used the same databases. Table~\ref{table:ComparasionOtherMethods} shows the results reported in the literature obtained on the IBSR and HFH datasets.

Liu et al.~\cite{liu2010fusing} have developed an auto context model
to segment the sub-cortical structures from T1W MRI. This
technique combines a discriminative model for appearance with a label
prior term. They have tested their approach on the IBSR database and
compared their results with FreeSurfer~\cite{fischl2002whole}, which
has been widely in this field. They reported average Dice coefficients
of $0.75$ and $0.74$ for the hippocampus in FreeSurfer and their
approach, respectively. The IBSR databased is also used for the
weighted voting method proposed by Artaechevarria et
al.~\cite{artaechevarria2009combination}. The best average Dice
coefficients were $0.74$ and $0.76$ for left and right hippocampus,
respectively. Rousseau et al.~\cite{rousseau2011supervised} proposed a
conventional patch-based labeling method using a fast multi-point
algorithm, i.e., only the voxels that are in the subdomain are
evaluated and a label patch is estimated in each voxel. These authors
have evaluated their implementation with the IBSR database. Their Dice
coefficients were $0.81$ and $0.81$ for left and right hippocampus, respectively. We report similar results for the conventional method. However, we demonstrate that our proposal has significant improvements over the conventional method.

Jafari-Khouzani et al.~\cite{jafari2011dataset} developed the HFH
database. They have evaluated two approaches on the HFH database: (i)
Parser~\cite{tu2008brain} and (ii) classifier fusion and labeling
(CFL)~\cite{aljabar2009multi}. Brain Parser uses Adaboost to select
and fuse a set of features from the training data to obtain the discriminative appearance model, which is combined with a generative shape model. Jafari-Khouzani et al. have reported average Dice coefficients of $0.64$ for the hippocampus. In CFL, the selected atlases are co-registered  to the target image, and their transferred labels are fused using the voting rule. The authors have reported Dice coefficients of $0.75$ for the hippocampus. Therefore, our results are as good as or even better than those previously reported.

\begin{table}
\begin{center}
\caption{Comparison of the proposed method with other segmentation
  methods using the mean DICE coefficient on the IBSR and HFH databases.}
\label{table:ComparasionOtherMethods}
\begin{tabular}{|l|c|c|c|c|c|} \hline
   Method & Proposed & Rousseau &  Fischl  & Liu  & Arteachavarria    \\
          &          &et al~\cite{rousseau2011supervised}&et al~\cite{fischl2002whole}&et al~\cite{liu2010fusing}&et al~\cite{artaechevarria2009combination}\\
   \hline
  LH - RH & 0.842-0.849 & 0.81-0.81 & 0.75 & 0.74 & 0.74-0.76 \\
   \hline
\end{tabular}

\begin{tabular}{|l|c|c|c|} \hline
   Method & Proposed & Brain Parser & CFL  \\
          &          &\cite{tu2008brain,jafari2011dataset} & \cite{aljabar2009multi,jafari2011dataset} \\
   \hline
  LH - RH & 0.790-0.804 & 0.64 & 0.75 \\
   \hline
\end{tabular}

\end{center}
\end{table}

%
\section{Discussion and conclusion}
In this work, we developed a patch-based labeling method that cooperates with atlas-warping using non-rigid registrations. First, a labeling of the target image is inferred with atlas-warping by non-rigid registrations. Then, a patch-based label fusion method is applied, whose patches and weights are computed from a combination of similarity measures between patches using intensity-based distances and labeling-based distances, where the labeling distances are calculated from the previous binary labeling of the target image by atlas-warping using non-rigid registrations.

The patch-based labeling methods have the advantages of considering
multiple samples during the labeling estimation and the local context
is well represented by the patches, particularly with affine
transformations. In contrast, the label fusion methods using non-rigid
registrations lead to segmentations with shape prior constraints. When
the delineation of the anatomical structures do not rely on intensity
contrast, as in the case of hippocampal segmentation, the conventional
patch-based labeling is not sufficient for obtaining good results. We have experimentally observed that the collaboration between these two approaches through the addition of a similarity measure based on the distance between binary labeling produces higher quality segmentations.

The collaboration between the two methods is given in the following
levels: (1) the sub-domain considered to accelerate the algorithm,
$\Omega^*$, generated by non-rigid registrations is smaller and has
more overlap with the manual segmentation than that obtained by affine
registrations. The consequence is a higher computational efficiency
due to the smaller size of $\Omega^*$ and improved segmentation
results by increasing the overlapping between the union of the
transferred labeled atlases and the ground-truth segmentations. (2)
The pre-selection of the patches in the atlases are improved by adding
similarity measures based on both the intensity and labeling of the
candidate patches. In our experiments, we observe that the numbers of
selected patches with our similarity measure are approximately
one-third less than that with the conventional method without
adding any additional parameter to tune. (3) The weights of our
selected patches are also more robust through the addition of
label-based distances. The segmentation results are the best compared
to other label fusion methods, including the conventional patch-based
labeling method. Moreover, our proposed method does not require
further tuning parameters compared to the conventional patch-based
methods, neither in the selected patches nor in calculating the
weights. (4) In the conventional patch-based labeling method, there
are no global constraints. Both in the pre-selection patch process and
in determining their weights, our approach imposes geometrical
restrictions, such as shape priors, using similarity measures based on
binary labeling. (5) We observe that an improvement in the
segmentation results using the label fusion method with non-rigid
registrations becomes an improvement of the proposed method. This is a
consequence of improvement in the selection and the robustness weights
of the patches. (6) The spatial normalization of the atlases and the
target images into a reference makes the work-flow very efficient. The
ranking of the atlases for both fusion methods and the library of the
patches are obtained from the spatial normalization. In the first
version of our sources, without the optimized code and only
parallelization of the non-rigid registration task and the patch-based
labeling, our proposal takes an average of approximately 3 and 6
minutes for segmenting the left and right hippocampus on images
belonging to the IBSR and HFH databases, respectively. Finally, we also
propose a type of patch adapted to the voxel spacing that provides
better results than standard solutions, which uses a cube shape. The scripts used in this study are available at \url{https://www.nitrc.org/projects/lf_patches/}.

In this paper, the patch-based labeling method is based on similarity measures of intensities and labeling. A new strategy of patch-based labeling has been proposed using the minimal reconstruction error~\cite{tong2013segmentation}. The patch library of the aligned atlases is considered as a dictionary, and the target patch is modeled as a sparse linear combination of the atoms in the dictionary. An extension of the patch reconstruction can be achieved by adding the inferred labeling from the label fusion using non-rigid registrations, in the same manner as performed on the similarity measures. Another limitation in this work is that the weights of the patches have been computed independently for each aligned atlas, without taking into account the fact that the selected atlases may produce similar label errors due to a high correlation between them. Wang et al.~\cite{wang2013multi} proposed that the weighted voting is formulated in terms of minimizing the total expectation of labeling error and in which pairwise dependency between atlases is explicitly modeled as the joint probability of two atlases, creating a segmentation error at a voxel. We leave these two themes (i.e., the minimal reconstruction error and the pairwise dependency between the patches of the atlases) for future work.










\end{document}